%% file: fully3d_template.tex
\documentclass[11pt,twocolumn,twoside]{article}
\usepackage{fully3d}
\input{math_commands.tex}

\begin{document}

\title{3D Photon Counting CT Image Super-Resolution Using Conditional Diffusion Model} 

\author[1$\dagger$]{Chuang Niu}
\author[1$\dagger$]{Christopher Wiedeman}
\author[1]{Mengzhou Li}
\author[2*]{Jonathan S Maltz}
\author[1*]{Ge Wang}

\affil[1]{Department of Biomedical Engineering, Center for Biotechnology \& Interdisciplinary Studies, Rensselaer Polytechnic Institute, Troy, NY USA}

\affil[2]{Molecular Imaging and Computed Tomography, GE HealthCare, Waukesha, WI, USA}

\affil[$\dagger$]{Co-first Authors}
\affil[*]{Co-corresponding Authors}

\maketitle
\thispagestyle{fancy}


\begin{customabstract}
This study aims to improve photon counting CT (PCCT) image resolution using denoising diffusion probabilistic models (DDPM).
Although DDPMs have shown superior performance when applied to various computer vision tasks, their effectiveness has yet to be translated to high-dimensional CT super-resolution.
To train DDPMs in a conditional sampling manner, we first leverage CatSim to simulate realistic lower-resolution PCCT images from high-resolution CT scans.
Since maximizing DDPM performance is time-consuming for both inference and training, especially on high-dimensional PCCT data, we explore both 2D and 3D networks for conditional DDPM and apply methods to accelerate training. In particular, we decompose the 3D task into efficient 2D DDPMs and design a joint 2D inference in the reverse diffusion process that synergizes 2D results of all three dimensions to make the final 3D prediction.
Experimental results show that our DDPM achieves improved results versus baseline reference models in recovering high-frequency structures, suggesting that a framework based on realistic simulation and DDPM shows promise for improving PCCT resolution.
\end{customabstract}


\vspace{-4mm}
\section{Introduction}
\vspace{-4mm}

Over the past two decades, CT imaging has rapidly advanced in terms of spatial, spectral and temporal resolution. However, for CT scanners based on conventional energy-integrating detectors (EIDs), further increases in spatial resolution are limited by the trade off that exists between reduced pixel size and dose-efficiency. This is because contemporary EIDs utilize segmented optical scintillators coupled to photodiode arrays. When pixel size is reduced, the reflective septa between pixels that prevent optical crosstalk occupy an increasingly fraction of the detector area, leading to losses in geometric- (and therefore dose-) efficiency.

Photon counting detectors (PCDs) are able to largely overcome this limitation. Photons incident on PCDs are directly converted into charge clouds in a semiconductor, and this charge is read out at electrodes. The nearest electrode to the point of photon interaction typically reads out the most charge, and so is identified as the pixel of interaction. The finer the spacing of the readout electrodes, the finer the potential spatial resolution of the detector. 

PCDs have indeed demonstrated the combination of higher spatial resolution and dose-efficiency than EID systems. However, several physical processes prevent PCDs from realizing resolution consistent with sampling provided by electrode spacing. (1) In reality, not all charge deposited in an interaction is collected by a single electrode, but is shared among nearby electrodes. (2) In high-Z detectors (such as CdTe/CZT), detector material K-edges are present within the energy range of the diagnostic energy X-ray spectrum. K-escape fluorescence thus occurs in these detectors, in which a photon deposits only a portion its energy at the point of initial interaction, and approximately 30~keV elsewhere. (3) In low-Z PCDs, such as edge-on-irradiated Si, Compton scatter occurs, so that some of the incident photon energy can leave a pixel \cite{danielsson2021photon}. (4) Some common PCD designs utilize a macropixel structure, with dead space between the macropixels to accommodate circuit elements such as through-silicon-bias. This leads to non-uniform sampling of the constituent micropixels, resolution loss, and aliasing artifact. All of these 4 processes and factors broaden the detector point-spread function in complex ways that depend on the incident spectrum, the spatial frequency content of the imaged object, and the material pathlengths traversed by each ray. Approaches such as anti-coincidence processing may be used to recover resolution for (1)--(3), but these tend to fail when flux is high due to the problem of pulse pileup. 

In this paper, our objective is to determine whether DL-based superresolution image postprocessing can superresolve photon counting CT images without implementing costly and potentially noise-enhancing deconvolutional methods based on physical processes that are too complex to model in practical imaging systems.

DL-based image superresolution (SR) has been extensively advanced in the computer vision field, including progress in terms of both network structure design (such as EDSR~\cite{lim2017enhanced}, SRGAN~\cite{ledig2017photo}, RCAN~\cite{zhang2018image}) and restoration frameworks (such as DPSR~\cite{zhang2019deep} and PULSE~\cite{menon2020pulse}). 
Despite the success of these techniques in the natural image domain, directly applying these methods to the medical image domain is challenging due to the lack of good quality low-resolution (LR) and high-resolution (HR) image pairs for network training. While downsampling techniques and  Gaussian noise models have been employed to generate synthetic datasets for CT image SR and achieve promising results~\cite{yu2017computed,you2019ct,jiang2020ct}, the metrics used to assess performance often do not translate to the desirability of the images from a clinical perspective. This is likely due to limitations of the image degradation model, as well as the quality metrics. For example, our recent study suggests that the insertion of unrealistically-distributed  noise can significantly degrade practical SR performance on images with real CT noise. Furthermore, the complex physics behind photon counting detection makes the degradation more challenging to represent realistically with simple analytic formulas~\cite{li2020x}. To address the challenge, this study aims to leverage advanced contemporary deep learning techniques and realistic simulation tools to improve PCCT resolution.

Recently, denoising diffusion probabilistic models (DDPM) \cite{ho2020denoising} have achieved great success in generative and reverse problems \cite{song2022solving,  chung2022improving, xia2022patch}. In comparison with adversarial generative models, DDPM does not suffer from mode-collapse and training instabilities, and demonstrates even better performance on various tasks. Nevertheless, effectively adapting the DDPM to improve the resolution of high-dimensional PCCT images has not yet been studied. A main obstacle we encountered in our initial application of DDPM to CT imaging is that directly training a DDPM as a conventional 3D network results in poor convergence and lengthy training times.
To overcome this challenge, we decompose the 3D task into two 2D models for improving in-plane and through-plane resolution respectively. However, the 2D model trained in one dimension usually exhibits degraded performance in other dimensions. To this end, we design a joint 2D inference in the reverse diffusion process that synergizes 2D results of all three dimensions to make the final 3D prediction.
We also design an alternative inference among different 2D models so that it is as efficient as a single 2D model inference.

Since it is not feasible to collect paired high and low resolution data that are perfectly registered, realistic simulation of aligned LR and HR is critical to building deep learning models. We use CatSim to generate low-resolution counterparts for CT image phantoms \cite{deman2007catsim, wu2022xcist}. Degradation is modulated by altering detector pitch, x-ray focal spot size, as well as noise and pixel cross-talk effects \emph{in silico}.

\vspace{-3mm}
\section{Methods}
\vspace{-3mm}
\subsection{Data and Simulation}
\def\LR{\textrm{\tiny{LR}}}
The CatSim PCCT module is to simulate scans of 10 digital phantoms. Each phantom is a reconstructed clinical CT head scan, which is converted into a water density voxel map based on attenuation. For proprietary reasons, absolute sinogram pixel size is suppressed; we denote the simulated LR pixel side lengths as $x_\LR$ and $z_\LR$. The LR and HR scans are simulated for 1000 views, $x_\LR \times z_\LR$ pixel size and 1~mm square focal spot; and 1300 views, $0.75\, x_\LR \times 0.85 \, z_\LR$ pixel size, and 0.75~mm focal spot, respectively. To achieve highest resolution and noise suppression, Poisson noise and pixel cross-talk are suppressed in HR scans. Simulating the phantoms at their true voxel size (0.293~mm in-plane, 0.625~mm axial) resulted in negligible difference between the phantom and LR reconstruction. We consequently reduced the voxel size by half in each direction to challenge the system.

All scans utilized 120~kVp tube voltage and 400~mA current for a one second rotation period and were reconstructed with filtered back projection such that there was a 1:1 voxel correspondence between the phantoms and reconstructions. One patient phantom (later used for testing) was scanned in parallel (i.e., superimposed in the sinogram domain) with in-plane and through-plane bar phantoms for quantitative evaluation.
\vspace{-3mm}
\subsection{Conditional DDPM} 
Following \cite{saharia2022image}, the CT SR task is formulated as a conditional generation. Given paired LR and HR images, $\{ \rvx_i, \, \rvy_i \}_{i=1}^N$ ($\rvx_i$ and $\rvy_i$ denote the LR and HR images, respectively, and $N$ is the number of image pairs) drawn from the conditional distribution $p(\rvy_i | \, \rvx_i)$, we aim to approximate $p(\rvy_i | \, \rvx_i)$ by learning a stochastic iterative process, where each and every iteration step is parameterized with the neural network function $f_{\rvtheta}$.
DDPM involves a forward Markovian process for training and a reverse Markovian diffusion process for inference.
Specifically, the forward process gradually adds Gaussian noise into an HR image via a fixed Markov chain, resulting in a series of images $\rvy_0 \rightarrow \rvy_1 \rightarrow \cdots \rightarrow \rvy_T$, where the noise level gradually increases with time step $t$, $\rvy_0$ and $\rvy_T$ are the HR and pure Gaussian noise image respectively, and $T$ is the number of iteration steps.
The forward Markovian diffusion process is defined by $q$:
\begin{align}
    q(\rvy_{1:T} | \, \rvy_0) &= \prod_{t=1}^T \, q(\rvy_t | \, \rvy_{t-1}), \\
    q(\rvy_{1:T} | \, \rvy_0) &= \bm{N}(\rvy_t | \sqrt{\alpha_t}\rvy_{t-1}, (1 - \alpha_t)\,\bm{I}),
\end{align}
where $\alpha_{\,1:T}$ are hyper-parameters that determine the variance of the Gaussian noise.
Fortunately, the distribution of $\rvy_t$ conditioned on $\rvy_0$ can be derived as:
\begin{equation}
    q(\rvy_t | \rvy_0) = \bm{N}(\rvy_t | \sqrt{\gamma_t}\, \rvy_0, (1 - \gamma_t)\bm{I}),
\end{equation}
where $\gamma_t = \prod_{i=1}^t \alpha_i$.
Thus, any intermediate noisy image $\rvy_t$ can be calculated given $\rvy_0$ as:
\begin{equation}
\label{eq_add}
    \rvy_t = \sqrt{\gamma_t}\,\rvy_0 + (1 - \gamma_t)\rvepsilon,  \ \ \ \rvepsilon \sim \bm{N}(\bm{0}, \bm{I}).
\end{equation}
The objective function used to training the network $f_{\rvtheta}$ to predict the Gaussian noise added in $\rvy_t$, conditioned on the LR image and the noise level $\gamma$ is:
\begin{equation}
\label{eq_pred}
    \E_{\rvx, \rvy} \, \E_{\rvepsilon, \gamma} \, || f_{\rvtheta}(\rvx, \sqrt{\gamma} \, \rvy_0 + \sqrt{1-\gamma} \, \rvepsilon, \gamma) - \rvepsilon ||_l^l,
\end{equation}
where we set $l=1$, $(\rvx, \rvy)$ is a paired training sample, $\gamma \sim p(\gamma)$ is defined as in \cite{saharia2022image}, and $\rvepsilon$ is normally-distributed noise. Given pure Gaussian noise $\rvy_T$ and LR image $\rvx$, the reverse Markovian process gradually removes noise from intermediate noisy images to generate the HR image, i.e., $\rvy_T \rightarrow \rvy_{T-1} \rightarrow \cdots \rightarrow \rvy_0$, which is defined by a parameterized distribution $p_{\rvtheta}$:
\begin{align}
    p_{\rvtheta}(\rvy_{0:T}) &= p_{\rvtheta}(\rvy_{T}) \prod_{t=1}^T \, p_{\theta}(\rvy_{t-1}\rvy_t, \rvx) \\
    p(\rvy_T) &= \bm{N}(\rvy_T | \bm{0}, \bm{I}) \\
    p_{\rvtheta}(\rvy_{t-1}| \rvy_t, \rvx) &= \bm{N}(\rvy_{t-1} | \, \mu_{\rvtheta}(\rvx, \rvy_t, \gamma_t), \sigma_t^2 \bm{I}),
\end{align}
where the reverse process $p(\rvy_{t-1} | \rvy_t, \rvx)$ is approximately Gaussian, assuming the noise variance in the forward steps is sufficiently small \cite{sohl2015deep}.
Based on Bayes' rule, the posterior distribution $\rvy_t$ conditioned on $(\rvy_0, \rvy_t)$ is:
\begin{align}
    &q(\rvy_{t-1} | \, \rvy_0, \rvy_t) = \bm{N}(\rvy_{t-1}| \, \bm{\mu}, \sigma^2\bm{I}) \\
    \label{eq_mu}
    &\bm{\mu} = \frac{\sqrt{\gamma_{t-1}} (1-\alpha_t)}{1 - \gamma_t} \, \rvy_0 + \frac{\sqrt{\alpha_t} (1-\gamma_{t-1})}{1 - \gamma_t} \, \rvy_t \\
    &\sigma^2 = \frac{(1-\gamma_{t-1})(1-\alpha_t)}{1-\gamma_t}.
\end{align}
Using Eqs. (\ref{eq_add}) and (\ref{eq_pred}), $\rvy_0$ can be approximated with the trained network as:
\begin{equation}
    \hat{\rvy}_0 = \frac{1}{\sqrt{\gamma_t}} \Big[\rvy_t - \sqrt{1-\gamma_t} f_{\theta}(\rvx, \rvy_t, \gamma_t)\Big].
\end{equation}
Replacing $\rvy_0$ in Eq. (\ref{eq_mu}) with $\hat{\rvy}_0$ yields:
\begin{equation}
    \mu_{\rvtheta}(\rvx, \rvy_t, \gamma_t) = \frac{1}{\sqrt{\alpha_t}} \Big[\rvy_t - \frac{1-\alpha_t}{\sqrt{1-\gamma_t}} f_\theta(\rvx, \rvy_t, \gamma_t)\Big],
\end{equation}
where the variance of $p_\rvtheta(\rvy_{t-1}|\,\rvy_t, \rvx)$ is $(1-\alpha_t)$ following \cite{ho2020denoising}.
Then, each iteration in inference is calculated as:
\begin{equation}
\label{eq_inf}
    \rvy_{t-1} \leftarrow \frac{1}{\sqrt{\alpha_t}} \Big[\rvy_t - \frac{1-\alpha_t}{\sqrt{1-\gamma_t}} f_\rvtheta(\rvx, \rvy_t, \gamma_t)\Big] + \sqrt{1 - \alpha_t}\, \rvepsilon_t.
\end{equation}
Thus, once the network is trained with Eq. (\ref{eq_pred}), the HR image can be iteratively generated from a pure Gaussian noise image conditioned on an LR image via Eq. (\ref{eq_inf}).

\vspace{-3mm}
\subsection{3D Super-Resolution with Joint 2D Inference}

We found that directly training a 3D version of the 2D network used in \cite{saharia2022image} on our CT datasets leads to poor convergence. In contrast, the corresponding 2D network can be easily trained.
A common strategy is to process the 3D CT scan slice-by-slice using a 2D network. However, this usually leads to degraded $z$-resolution. To overcome this problem, we propose to train two 2D networks to improve in-plane and through-plane resolution respectively, where the coronal and sagittal planes share the same though-plane network.
To improve resolution in three dimensions, we constrain the results of all dimensions to be consistent in the inference stage. A simple implementation is to find a consistent 3D sampling at each time step, i.e., $ \Tilde{\rvy}_t = \argmin_{\rvy_t}(\lambda_c||\, \rvy_t - \rvy_t^h||_2^2 + \lambda_c||\, \rvy_t - \rvy_t^c||_2^2 + \lambda_h||\, \rvy_t - \rvy_t^s||_2^2) = \frac{1}{\lambda_h+\lambda_c + \lambda_s} (\lambda_h \rvy_t^h + \lambda_c \rvy_t^c + \lambda_s \rvy_t^s)$ and then use $\Tilde{\rvy}_t$ as the input for next iteration, where $\rvy_t^h$, $\rvy_t^c$, and $\rvy_t^s$ are the 2D prediction results for the three dimensions.
However, this implementation increases inference time threefold.
Instead, we propose an efficient strategy for merging 2D results, which alternately performs a single 2D inference for one of three dimensions at each step, using each 2D result as the input for the next iteration. Computational cost is not increased, as each iteration only requires a single 2D network. In practice, these two implementations yield comparable results.
\vspace{-3mm}
\subsection{Implementation Details}
We use 9 patient CT scans for training and 1 patient CT scan for testing. 128 $\times$ 128 patches are randomly cropped for training the 2D networks. We use the same network architecture as in \cite{saharia2022image}, and the attention is applied to the layer with the smallest spatial dimension. The batch size is 4 and the Adam optimizer uses a learning rate of $10^{-4}$. The number of sampling steps for DDPM is set to 2000, the number of training iterations to 300000, and all other hyper-parameters are set equal to those employed in \cite{saharia2022image}.
To compare DDPM models with a conventional supervised learning method, we modify the 2D network to its 3D version by converting all 2D operations to 3D ones. Since the 3D network significantly increases the memory cost, we reduce the number of inner channels to fit GPU devices, e.g., for a 24~GB GPU with a single $128\times 128\times 128$ sub-volume, the number of base inner channels is set to 12, and the channel multipliers are 1, 2, 4, 8, 8 for the five blocks, respectively. All other hyper-parameters are identical to those used for 2D DDPMs.
Our implementation is based on PyTorch, and it is well-known that automatic mixed precision will significantly improve the training speed. However, we find this sometimes makes the training process unstable in our experiments, so this technique is not used in this study.
\vspace{-4mm}
\section{Experiments and Results}
\begin{figure}[bt!]
    \centering
    \includegraphics[width=0.4\textwidth]{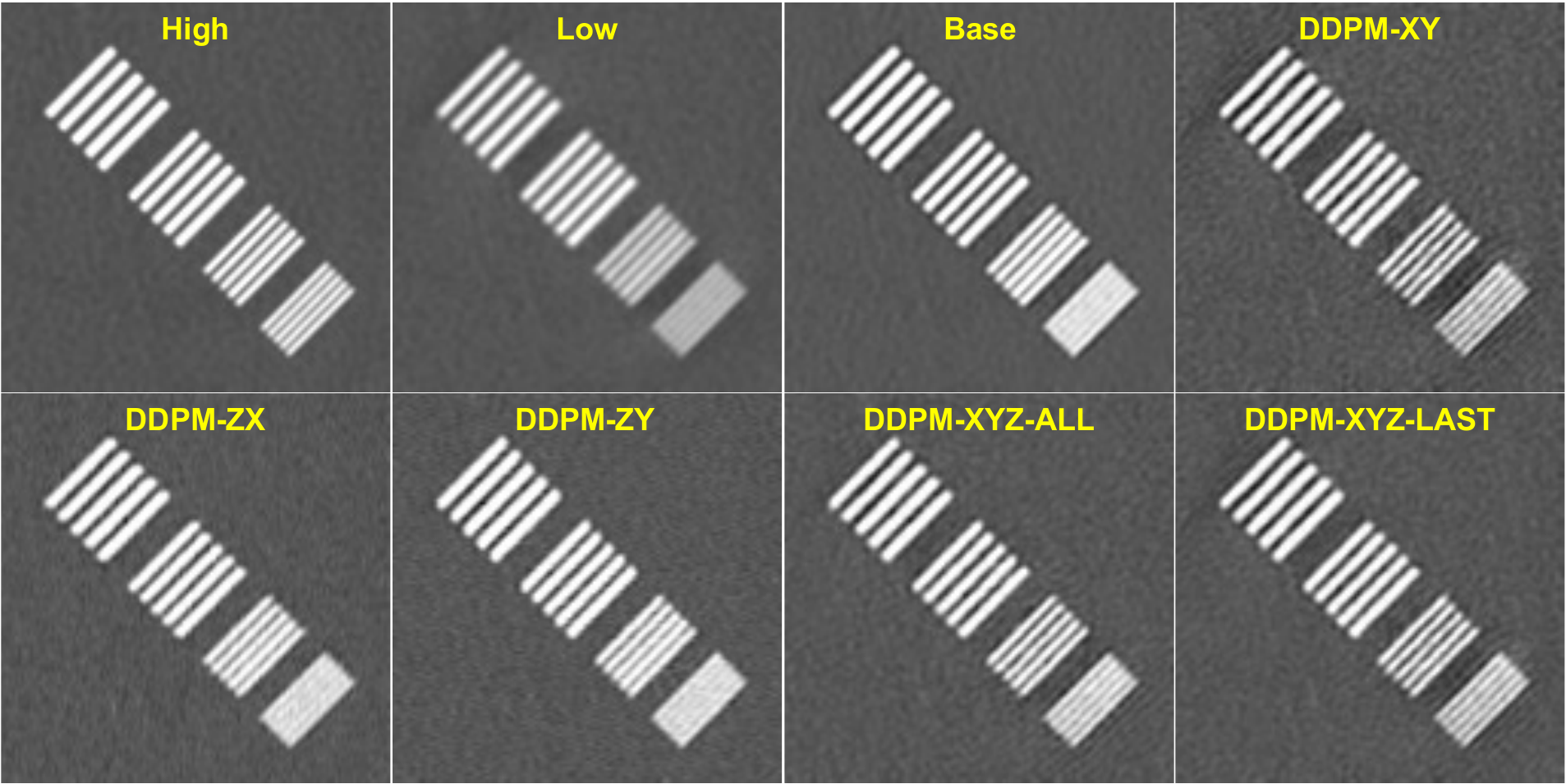}
    \caption{In-plane (XY) results.
    }
    \label{fig_bxy}
\end{figure}

\begin{figure}[bt!]
    \centering
    \includegraphics[width=0.4\textwidth]{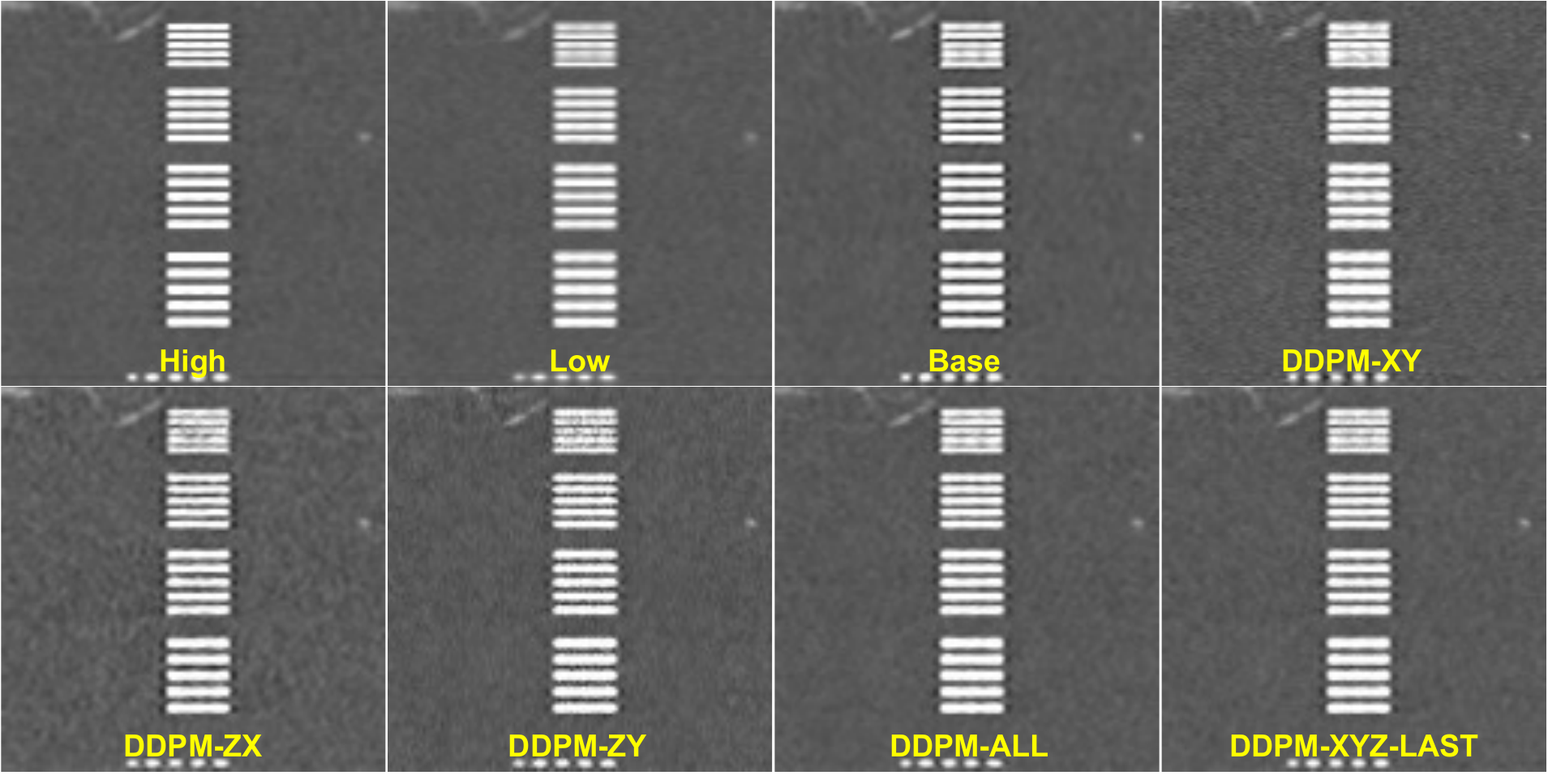}
    \caption{Through-plane ($z$--$x$) results.
    }
    \label{fig_bxz}
\end{figure}

\begin{figure}[bt!]
    \centering
    \includegraphics[width=0.5\textwidth]{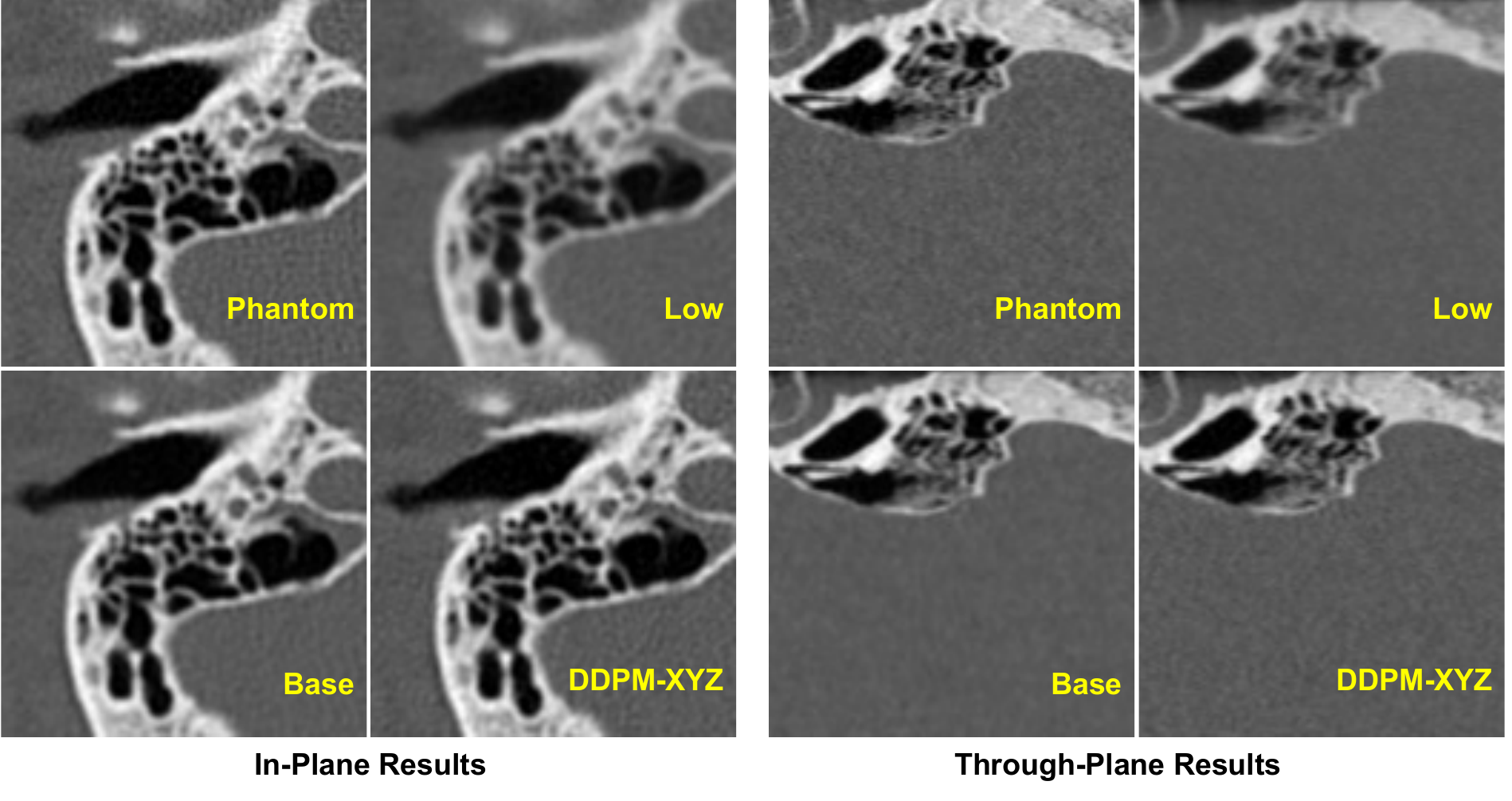}
    \caption{Application to CT images of the human temporal bone.
    }
    \label{fig_rs}
\end{figure}

\vspace{-4mm}
In our pilot experiments, we evaluate different variants of DDPMs and the baseline 3D model trained with conventional supervised learning on the test patient CT scan with the inserted line pairs. Note that the line pair patterns are not included in the training dataset.
Here we evaluate two variants of 2D joint inference: 1) DDPM-XYZ-ALL alternatively does 2D inference to merge 2D results among all inference steps; 2) DDPM-XYZ-LAST calculates the weighted sum of 2D results that are independently computed in the last step only, increasing the inference time threefold. 
The in-plane and through-plane results are shown in Figures \ref{fig_bxy} and \ref{fig_bxz} respectively.
We observe (1) the 2D DDPM trained with in-plane slices visibly improves the resolution of in-plane slices, but its performance on through-plane dimensions is degraded. Also, some line artifacts are generated around the high-frequency line pairs. Generally, the 2D network trained on a specific dimension does not work well on other dimensions. 
(2) The joint inference that synergizes all-dimension results can clearly improve all-dimension performance in certain respects, e.g., artifacts present in the 2D in-plane results are reduced.
(3) DDPMs appear to achieve better super-resolution results than the baseline models, recovering more detail. We also evaluated the presented DDPM and baseline models on anatomical structures. The results in Figure \ref{fig_rs} show that, in comparison with the baseline model, DDPM achieves sharper results and the image texture better resembles that of the high-resolution phantoms for both in-plane and through-plane slices.

\begin{figure}[bt!]
    \centering
    \includegraphics[width=0.4\textwidth]{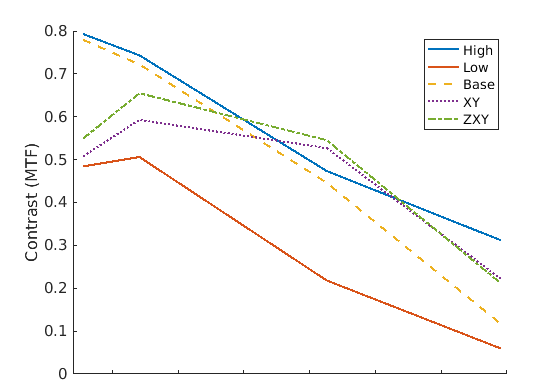}
    
    \caption{MTF Comparison (In-Plane). The frequency axis has been suppressed for proprietary reasons.
    }
    \label{fig_mtfxy}
\end{figure}

\begin{figure}[bt!]
    \centering
    \includegraphics[width=0.4\textwidth]{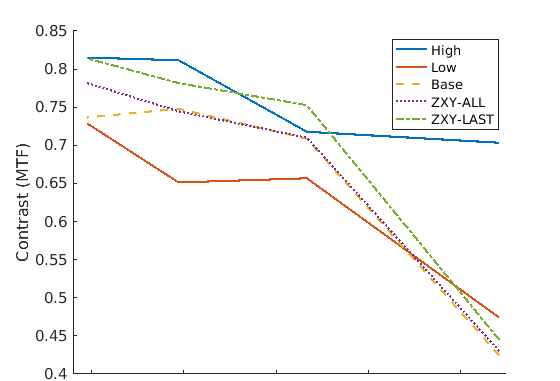}
    \caption{MTF Comparison (Through-Plane).  The frequency axis has been suppressed for proprietary reasons.
    }
    \label{fig_mtfz}
\end{figure}

To quantitatively evaluate the resolution results, we calculate the corresponding modulation transfer function (MTF) of the line pairs in Figures \ref{fig_bxy} and \ref{fig_bxz}, and the results are shown in Figures \ref{fig_mtfxy} and \ref{fig_mtfz} correspondingly, where the best two DDPM results are displayed.

\vspace{-5mm}
\section{Discussion and conclusion}
\vspace{-3mm}
Figure \ref{fig_mtfxy} suggests that the in-plane resolution is improved by the DDPMs; this improvement is superior to that of the baseline model at higher frequencies, even rivaling the high-resolution reference. MTF comparison for axial resolution is slightly more ambiguous, but the select DDPMs still generally outperform the baseline model (Figure \ref{fig_mtfz}). It should be noted that MTF measures resolution by contrast in a sinusoidal pattern, which may not comprehensively evaluate edge sharpness. Nevertheless, these quantitative results, combined with the previous qualitative observations, support the potential of DDPMs in CT super-resolution.

We have demonstrated the effectiveness of conditional DDPMs in the PCCT super-resolution task. We have overcome a major challenge of training high-dimensional DDPMs by training in-plane and through-plane 2D networks, and then synergizing the 2D predictions of all dimensions. Experimental results have demonstrated the effectiveness of the presented method.

\vspace{-3mm}
\printbibliography

\end{document}

%% file: math_commands.tex
\usepackage{amsmath,amsfonts,bm}

\def\1{\bm{1}}

\def\rvepsilon{{\mathbf{\epsilon}}}
\def\rvtheta{{\mathbf{\theta}}}

\def\rvx{{\mathbf{x}}}
\def\rvy{{\mathbf{y}}}

\DeclareMathAlphabet{\mathsfit}{\encodingdefault}{\sfdefault}{m}{sl}
\SetMathAlphabet{\mathsfit}{bold}{\encodingdefault}{\sfdefault}{bx}{n}

\newcommand{\E}{\mathbb{E}}

\DeclareMathOperator*{\argmin}{arg\,min}